\documentclass[letterpaper,twocolumn,10pt]{article}
\usepackage{usenix2019_v3}
\usepackage{tikz}
\usepackage{enumitem}
\usepackage{xspace}

\newcommand{\websiteurl}{\url{dlspec.netlify.com}\xspace}

\DeclareRobustCommand*\circledwhite[1]{\tikz[baseline=(char.base)]{
            \node[shape=circle,line width=0.2mm,draw,inner sep=0.5pt,anchor=base] (char) {\normalfont\sffamily\bfseries\footnotesize{#1}};}}

\begin{document}

\date{}
\title{\Large \bf DLSpec: A Deep Learning Task Exchange Specification}

\author{{\rm Abdul Dakkak\textsuperscript{1*}, Cheng Li\textsuperscript{1*}, Jinjun Xiong\textsuperscript{2}, Wen-mei Hwu\textsuperscript{1}} \\ \textsuperscript{1}University of Illinois Urbana-Champaign, \textsuperscript{2}IBM T. J. Watson Research Center \\
\{dakkak, cli99, w-hwu\}@illinois.edu, jinjun@us.ibm.com
}

\maketitle

\renewcommand{\thefootnote}{\fnsymbol{footnote}}
\footnotetext[1]{The two authors contributed equally to this paper.}

\begin{abstract}

Deep Learning (DL) innovations are being introduced at a rapid pace.
However, the current lack of standard specification of DL tasks makes sharing, running, reproducing, and comparing these innovations difficult.
 To address this problem, we propose DLSpec, a model-, dataset-, software-, and hardware-agnostic DL  specification that captures the different aspects of DL tasks. 
 DLSpec has been tested by specifying and running hundreds of DL tasks.
\end{abstract}

\vspace{-10pt}
\section{Introduction}\label{sec:intro}
\vspace{-5pt}

The past few years  have seen a fast growth in Deep Learning (DL) innovations such as datasets, models, frameworks, software, and hardware.
Current practice of publishing these DL innovations involves developing ad-hoc scripts and writing textual documentation to describe the execution process of \textit{DL tasks} (e.g. model training or inference).
This requires a lot of effort and  makes sharing and running DL 
tasks  
 difficult.
Moreover, it is often hard to reproduce reported accuracy or performance results and have a consistent comparison across DL tasks.
This is a known~\cite{gundersen2018state,hutson2018artificial}  ``pain point’’ within the DL community.
Having an exchange specification to describe DL tasks would be a first step to remedy this and ease the adoption of DL innovations.

Previous work included curation of DL tasks in framework model zoos~\cite{modelhub, modelzoo, caffe2zoo,gluoncv,onnxzoo,tensorflowhub},  developing model catalogs that can be used through a cloud provider's API~\cite{gcp,azure,sagemaker}, or introducing MLOps specifications~\cite{vartak2016modeldb,zaharia2018accelerating,fursin2020codereef}. 
However, these work 
either use ad-hoc techniques for different DL tasks or are tied to a specific hardware or software stack.

We propose DLSpec, a DL artifact exchange specification 
with clearly defined model, data, software, and hardware aspects.
DLSpec's design is based on a few key principles (Section~\ref{sec:objectives}).
DLSpec is model-, dataset-, software-, and hardware-agnostic and 
aims to work with runtimes built using existing MLOp tools.
We further develop a DLSpec runtime  to support DLSpec's use for DL inference tasks in the context of benchmarking~\cite{li2019design}.

\begin{figure*}
  \centering
  \vspace{-25pt}
  \setlength{\abovecaptionskip}{0pt}
  \includegraphics[clip, width=0.9\textwidth]{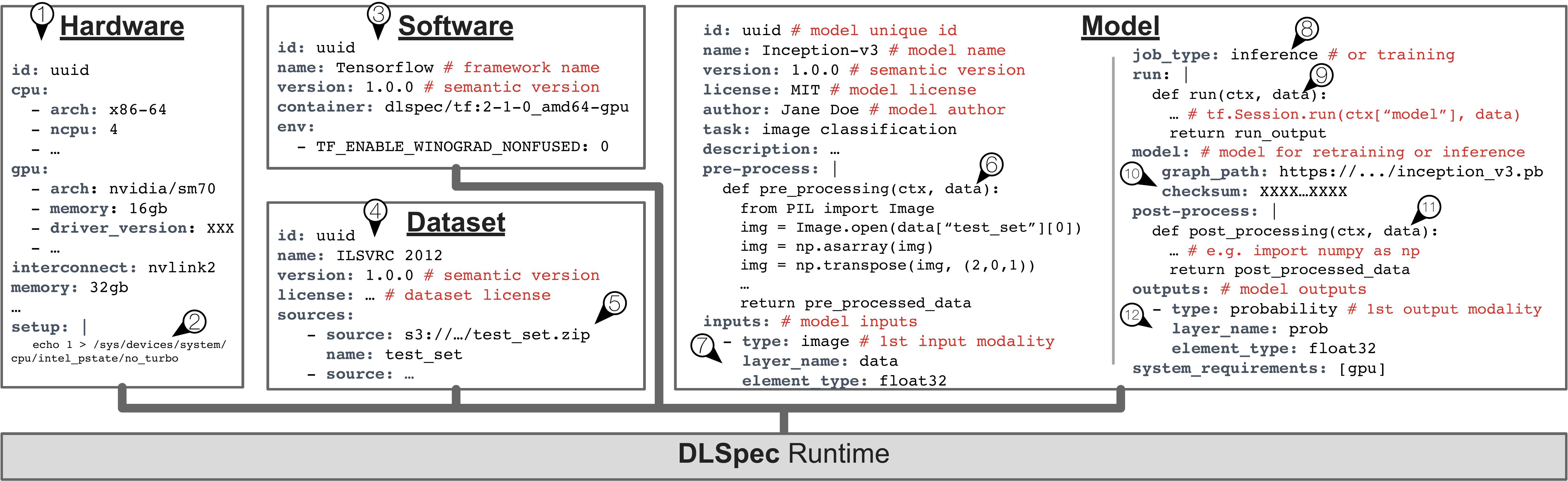}
  \caption{An example DLSpec that consists of a hardware, software, dataset and model manifest. }
  \vspace{-15pt}
  \label{fig:spec}
\end{figure*}

\section{Design Principles}\label{sec:objectives}
\vspace{-8pt}

While the bottom line of a 
specification design is to ensure the usability and reproducibility of DL tasks, %
 the following principles 
are considered to increase 
DLSpec's applicability: %

\noindent
\textbf{Minimal} ---
To increase the transparency and ease the creation, the specification 
should contain only the essential information to use a task and reproduce %
its reported outcome.

\noindent
\textbf{Program-/human-readable} ---
To make it possible to develop a runtime that executes DLSpec, the specification should be readable by a program.
To allow a user to understand what the task does and repurpose it (e.g. use a different HW/SW stack), the specification should be easy to introspect.

\noindent
\textbf{Maximum expressiveness} ---
While DL training and inference tasks can share many common software and hardware setups, there are differences when specifying their resources, parameters, inputs/outputs, metrics, etc.
The specification must be general to be used for both training and inference tasks.

\noindent
\textbf{Decoupling DL task
description} ---
A DL 
task 
is described from the aspects of model, data, software, and hardware through their respective manifest files.
Such a decoupling increases the reuse of manifests and enables the portability of DL tasks across datasets, software, and hardware stacks. 
This further enables one to easily compare different DL offerings by varying one of the four aspects.

\noindent
\textbf{Splitting the DL task pipeline stages} ---
We demarcate the stages of a DL task into pre-processing, run, and post-processing stages. This enables consistent comparison and simplifies accuracy and performance debugging. 
For example, to debug accuracy, one can modify the pre- or post-processing step and observe the accuracy; and 
to debug performance, one can surround the run stage with the measurement code.
This demarcation is consistent with existing best practices~\cite{mattson2019mlperf,reddi2019mlperf}.

\noindent
\textbf{Avoiding serializing intermediate data into files} --- 
A naive way to transfer data between stages of a DL task is to use files.
In this way, each stage reads a file containing input data and writes to a file with its output data.
This approach can be impractical since it introduces high serializing/deserializing overhead. %
Moreover, this technique would not be able to support DL tasks that use streaming data.

\vspace{-13pt}
\section{DLSpec Design}\label{sec:design}
\vspace{-8pt}

A DLSpec consists of four manifest files and a reference log file.
All manifests are versioned~\cite{semanticver} and have an ID (i.e. can be referenced).
The four manifests (shown in Figure~\ref{fig:spec}) are:

\begin{itemize}[nosep,leftmargin=0.0em,labelwidth=*,align=left]
    \item \textbf{Hardware}: defines the hardware requirements for  a DL task. The parameters form a series of hardware
    constraints that must be satisfied to run and reproduce a task. %
    \item \textbf{Software}: defines the software environment for a DL task. All executions occur within the specified container.
    \item \textbf{Dataset}: defines the training, validation, or test dataset.
    \item \textbf{Model}: defines the logic (or code) to run a DL task and the required artifact sources.
\end{itemize}

The reference log is provided by the specification author for others to refer to.
The reference log contains the IDs of the manifests used to create it, achieved accuracy/performance of DL task, expected outputs, and author-specified information (e.g. hyper-parameters used in training).

We highlight key details of DLSpec:

\noindent
\textbf{Containers} ---
A software manifest specifies the container to use for a DL task, as well as any configuration parameters that must be set (e.g. framework environment variables).
The framework or other libraries information are listed for ease of inspection and management.

\noindent
\textbf{Hardware configuration} --- 
While containers provide a standard way of specifying the software stack, a user cannot specify some hardware settings within a container.
E.g., it is not possible to turn off Intel's turbo-boosting (Figure~\ref{fig:spec}\circledwhite{2}) within a container.
Thus, DLSpec specifies hardware configurations in the hardware manifest to allow the runtime to set them outside the container environment.

\noindent
\textbf{Pre-processing, run, and post-processing stages} ---
The pre-/post-processing and run stages are defined via Python functions embedded within the manifest.
We do this because a DLSpec runtime can use the Python sub-interpreter~\cite{python-subinterpreter} to execute the Python code within the process thus avoiding using intermediate files (see Section~\ref{sec:objectives}).
Using Python functions also allows for great flexibility; e.g. the Python function can download and run Bash and R scripts or download, compile, and some C++ code.
The signature of the DLSpec Python functions is \texttt{fun(ctx, data)} where \texttt{ctx} is a hash table that includes manifest information (such as the types of inputs) accepted by the model.
The second argument, \texttt{data}, is the output of the previous step in the dataset$\rightarrow$pre-processing$\rightarrow$run$\rightarrow$post-processing pipeline.
In  Figure~\ref{fig:spec}, for example, the pre-processing stage's \circledwhite{6} \texttt{data} is the list of file paths of the input dataset (ImageNet test set in this case).

\noindent
\textbf{Artifact resources} ---
DL artifacts used in a DL task are specified as remote resources within DLSpec.
The remote resource can be hosted on an FTP, HTTP, or file server (e.g. AWS S3, Zenodo) and have a checksum which is used to verify the download.

\vspace{-13pt}
\section{DLSpec Runtime}\label{sec:runtime}
\vspace{-8pt}

While a DL practitioner can run a DL task by manually following the setup described in the manifests, here we describe how a runtime (i.e. an MLOps tool) can use the DLSpec manifests shown in Figure~\ref{fig:spec}.

A DLSpec runtime consumes the four manifests and selects the \circledwhite{1} hardware to use and runs any \circledwhite{2} setup code specified (outside the container).
A \circledwhite{3} container is launched using the image specified, and the  \circledwhite{4} dataset is downloaded into the container using the \circledwhite{5} URLs provided.
The  \circledwhite{6} dataset file paths are passed to the pre-processing function and its outputs are then processed to match the \circledwhite{7}  model's input parameters.
The \circledwhite{9} DL task is run.
In the case of \circledwhite{8} inference, this causes the \circledwhite{10} model to be downloaded into the container.
The result from the run are then \circledwhite{11} post-processed using the \circledwhite{12} data specified by the model outputs.

We  tested DLSpect
in the context of inference benchmarking and implemented a runtime for it~\cite{li2019design}.
We 
collected over 300 popular models and created  reusable 
manifests for each.
We created software manifests for major frameworks (Caffe, Caffe2, CNTK, MXNet, PyTorch, TensorFlow, TensorFlow Lite, and TensorRT), dataset manifests (for ImageNet, COCO, Pascal, CIFAR, etc.), and then wrote hardware specs for X86, ARM, and  PowerPC.
We tested our design and showed that it enables consistent and reproducible evaluation of DL 
tasks 
at scale.

\vspace{-13pt}
\section{Conclusion}\label{sec:conclusion}
\vspace{-8pt}

An exchange specification, such as DLSpec, enables a streamlined way to share, reproduce, and compare DL tasks.
DLSpec takes the first step in defining a DL task for both training and inference and captures the different aspects of DL model reproducibility.
We are actively working on refining the specifications as new DL tasks are introduced.
We maintain an updated published version of DLSpec at \websiteurl{}.

\bibliographystyle{plain}
\bibliography{main}

\end{document}